\documentclass{article}

\usepackage[acronym,nowarn,section,nogroupskip,nonumberlist]{glossaries}

\glsdisablehyper{}
\newcommand{\acaps}[1]{{\scshape #1}}

\newacronym{gnn}{\acaps{gnn}}{Graph Neural Network}
\newacronym{ssl}{\acaps{ssl}}{Self-Supervised Learning}
\newacronym{fc}{\acaps{fc}}{Functional Connectivity}
\newacronym{fmri}{f\acaps{mri}}{functional Magnetic Resonance Imaging}
\newacronym{roi}{\acaps{roi}}{Regions of Interest}
\newacronym{vae}{\acaps{vae}}{Variational Auto Encoder}
\newacronym{sce}{\acaps{sce}}{Scaled Cosine Error}
\newacronym{bold}{\acaps{bold}}{Blood-Oxygen-Level-Dependent}
\newacronym{gru}{\acaps{gru}}{Gated Recurrent Unit}
\newacronym{bce}{\acaps{bce}}{Binary Cross-Entropy}
\newacronym{gin}{\acaps{gin}}{Graph Isomorphism Network}
\newacronym{auroc}{\acaps{auroc}}{Area Under the Receiver Operating Characteristics}
\newacronym{mae}{\acaps{mae}}{Mean Absolute Error}
\newacronym{gcl}{\acaps{gcl}}{Graph Contrastive Learning}
\newacronym{mse}{\acaps{mse}}{Mean Squared Error}
\newacronym{sce2}{\acaps{sce}}{Softmax Cross-Entropy}
\newacronym{mlp}{\acaps{mlp}}{Multi-Layer Perceptron}
\usepackage{amsmath, amssymb, amsthm}
\usepackage{mathtools}
\usepackage{bbm}
\usepackage{dsfont}

\newcommand{\bW}{\mathbf{W}}

\newcommand{\bse}{\boldsymbol{e}}

\newcommand{\bsv}{\boldsymbol{v}}

\newcommand{\bsx}{\boldsymbol{x}}

\newcommand{\bsA}{\boldsymbol{A}}

\newcommand{\bsH}{\boldsymbol{H}}

\newcommand{\bsP}{\boldsymbol{P}}

\newcommand{\bsW}{\boldsymbol{W}}
\newcommand{\bsX}{\boldsymbol{X}}

\newcommand{\bsZ}{\boldsymbol{Z}}

\newcommand{\calE}{{\mathcal{E}}}
\newcommand{\calF}{{\mathcal{F}}}

\newcommand{\calL}{{\mathcal{L}}}

\newcommand{\calS}{{\mathcal{S}}}
\newcommand{\calT}{{\mathcal{T}}}

\newcommand{\calV}{{\mathcal{V}}}

\newcommand{\bbR}{\mathbb{R}}

\newcommand{\boldeta}{{\boldsymbol{\eta}}}

\theoremstyle{plain}%

\theoremstyle{definition}

\theoremstyle{remark}

\def\[#1\]{\begin{equation}\begin{aligned}#1\end{aligned}\end{equation}}

\usepackage[usenames,dvipsnames]{xcolor}

\usepackage{graphicx}

\usepackage{booktabs, array}
\usepackage{multirow}
\usepackage{colortbl}
\definecolor{LightCyan}{rgb}{0.88,0.95,1.0}

\newsavebox\CBox

\definecolor{citecolor}{RGB}{0,102,204}
\definecolor{linkcolor}{RGB}{190,105,30}
\definecolor{urlcolor}{RGB}{199,21,133}
\definecolor{blackcolor}{RGB}{0,0,0}

\usepackage[colorlinks, linktoc=all ,pagebackref=true]{hyperref}
\usepackage[all]{hypcap}
\hypersetup{citecolor=citecolor}
\hypersetup{linkcolor=linkcolor}
\hypersetup{urlcolor=urlcolor}
\usepackage[nameinlink, capitalise, noabbrev]{cleveref}
\creflabelformat{equation}{#2\textup{#1}#3}  %
\crefname{section}{\S}{\S\S}

\usepackage[utf8]{inputenc} %
\usepackage[T1]{fontenc}    %
\usepackage{hyperref}       %
\usepackage{url}            %
\usepackage{booktabs}       %
\usepackage{amsfonts}       %
\usepackage{nicefrac}       %
\usepackage{microtype}      %
\usepackage{xcolor}         %
\usepackage{setspace}

\usepackage{multirow}
\usepackage{adjustbox}
\usepackage{caption}
\usepackage{booktabs}
\usepackage{pifont} %
\usepackage{algorithm}
\usepackage{algorithmic}

\usepackage{subcaption}
\usepackage{duckuments}
\usepackage{booktabs}
\usepackage{arydshln}
\usepackage[dvipsnames]{xcolor}
\usepackage{color, colortbl}

\usepackage{amssymb}%
\usepackage{pifont}%
\usepackage{wrapfig}

\usepackage[numbers]{natbib}
\usepackage[final]{neurips_2023}

\title{A Generative Self-Supervised Framework using Functional Connectivity in fMRI Data}

\author{
  Jungwon Choi$^{1}$, Seongho Keum$^{1}$, EungGu Yun\thanks{Independent researcher / $^\dagger$Corresponding author.}\,\,\,
  , Byung-Hoon~Kim$^{\dagger\,2\,3}$, Juho Lee$^{\dagger\,1\,4}$ \\
  $^{1}$KAIST AI, $^{2}$Yonsei University College of Medicine, 
  $^{3}$MGH, Harvard Medical School, $^4$AITRICS \\
  \texttt{\{jungwon.choi, shkeum, eunggu.yun\}@kaist.ac.kr}, \\
  \texttt{egyptdj@yonsei.ac.kr}, \texttt{juholee@kaist.ac.kr}
}

\begin{document}

\maketitle
\begin{abstract}
Deep neural networks trained on \gls{fc} networks extracted from \gls{fmri} data have gained popularity due to the increasing availability of data and advances in model architectures, including \gls{gnn}. 
Recent research on the application of \gls{gnn} to \gls{fc} suggests that exploiting the time-varying properties of the \gls{fc} could significantly improve the accuracy and interpretability of the model prediction. 
However, the high cost of acquiring high-quality \gls{fmri} data and corresponding phenotypic labels poses a hurdle to their application in real-world settings, such that
a model naïvely trained in a supervised fashion can suffer from insufficient performance or a lack of generalization on a small number of data.
In addition, most \gls{ssl} approaches for \glspl{gnn} to date adopt a \emph{contrastive} strategy, which tends to lose appropriate semantic information when the graph structure is perturbed or does not leverage both spatial and temporal information simultaneously.
In light of these challenges,
we propose a \emph{generative} \gls{ssl} approach that is tailored to effectively harness spatio-temporal information within dynamic \gls{fc}. 
Our empirical results, experimented with large-scale (>50,000) \gls{fmri} datasets, demonstrate that our approach learns valuable representations and enables the construction of accurate and robust models when fine-tuned for 
downstream tasks.
\end{abstract}
\section{Introduction}
\label{main:sec:introduction}

The investigation into the complexities of human brain functionality has seen significant strides with the advent of neuro-imaging techniques \citep{ktena2018metric}. Among these, \gls{fmri} is considered a pivotal modality. It captures \gls{bold} signals, offering an in-depth view of the brain's neural activity with relatively high spatial and temporal resolution. 
Leveraging \gls{fc} based on \gls{fmri} data has become increasingly popular in solving a myriad of problems related to the human brain
\citep{arslan2018graph, kim2020understanding}. 
\gls{fc} allows the formation of graphs that represent connections between \glspl{roi} in the brain, thereby transforming the problem into a graph-learning task.

To add to the complexity, acquiring labeled \gls{fmri} data is an expensive and laborious process, often resulting in limited availability of labeled data for supervised learning \citep{miller2016multimodal, alfaro2018image}. This challenge is not unique to \gls{fmri} but is a common hurdle in many real-world applications such as fraud detection, event forecasting, and recommendation systems. \gls{ssl} thus appears as a compelling solution to leverage the plethora of unlabeled \gls{fmri} data to learn useful features for downstream tasks \citep{chen2020simple, devlin2018bert, you2020graph}.

However, most existing \gls{ssl} approaches for graph data, including \gls{fc} networks, focus solely on static graphs, ignoring the temporal dynamics that are often crucial for understanding complex systems \citep{xia2022simgrace, yu2022graph, hou2022graphmae, tan2022mgae, li2023s, li2023augmentation, hou2023graphmae2}. This is a significant limitation, as many real-world networks, including brain networks, social networks, and financial systems, are inherently dynamic. They evolve over time, and this temporal information can be crucial for various applications like anomaly detection and recommendation systems.

To address this gap, we introduce a novel framework named Spatio-Temporal Masked Auto-Encoder (ST-MAE) specifically tailored for \gls{fmri} data. Unlike conventional methods that mask nodes or edges in static graphs, ST-MAE learns node representations that capture the temporal knowledge inherent in dynamic graphs. Specifically, ST-MAE employs representations from different time stamps to reconstruct masked node features at intermediate time stamps. We pre-train our model on a large-scale UKB~\citep{sudlow2015uk} dataset, comprising approximately 40,000 entries, transforming it into \gls{fc}-based dynamic graphs. Our methodology undergoes extensive validation against various benchmarks including ABCD~\citep{casey2018adolescent}, HCP~\citep{wu20171200}, HCP-A~\citep{bookheimer2019lifespan}, HCP-D~\citep{somerville2018lifespan}, ABIDE~\citep{craddock2013neuro}, and ADHD200~\citep{brown2012adhd}. The results demonstrate a notable improvement in downstream fMRI tasks.

The primary contributions of our work are as follows:
\begin{itemize}
    \item We are the first to propose a Generative \gls{ssl} framework for dynamic graphs that takes into account temporal features for pre-training, introducing the concept of Spatio-Temporal Masked Auto-Encoder (ST-MAE).
    \item We utilize the large-scale UKB dataset to create \gls{fc}-based dynamic graphs and demonstrate the capability of \gls{ssl} in capturing meaningful \gls{fmri} representations for downstream tasks.
    \item Our framework excels particularly in the classification of psychiatric disorders, highlighting its utility in scenarios with limited labeled data.
\end{itemize}
\vspace{-3mm}

\section{Background}
\label{main:sec:background}

\vspace{-1mm}
\subsection{Settings and Notations for Dynamic Graphs}
\vspace{-1mm}

A static graph $ \mathcal{G} = (\mathcal{V}, \mathcal{E}) $ consists of a vertex set $ \mathcal{V} $ and an edge set $ \mathcal{E} $. In contrast, a dynamic graph $ G_{\text{dyn}} $ is defined as a sequence of graphs $ \mathcal{G}(t) $ at discrete time points $ t $. Each $ \mathcal{G}(t) $ is described by an adjacency matrix $ A(t) $ and node feature vectors $ \bsx_v(t) $ where $\bsv \in \mathcal{V}$. Formally, a dynamic graph $ G_{\text{dyn}} $ can be defined as:
\vspace{-5mm}

\begin{equation}
G_{\text{dyn}} = \{ \mathcal{G}(1), \mathcal{G}(2), \ldots, \mathcal{G}(T) \}, \quad
\bsA(t) = [a_{ij}(t)] \in \{0, 1\}^{N \times N},
\end{equation}
\vspace{-5mm}

where the number of nodes $N$ is assumed to be fixed throughout time and $T$ represents the total number of timepoints in the dynamic graph. In order to capture the temporal variations in node features, we employ a time encoding vector $ \boldeta(t) \in \bbR^D $, which can be generated using a sequence model such as \gls{gru} following \citet{kim2021learning}, where $D$ is the size of hidden dimension. The final node feature vector at time $t$ is then defined as $\bsx_v(t) = \bsW[\bse_v \Vert \boldeta(t)]$ where $\bsW \in \bbR^{(N+D) \times D}$ is a learnable matrix, $\bse_v \in \bbR^{N \times N}$ is the spatial feature encoding of the node, $\boldeta(t)$ is the temporal feature encoding, and $\Vert$ is a concatenation operation.
\vspace{-1mm}

\subsection{Masked Autoencoders in Static Graph}
\vspace{-1mm}

A Masked Autoencoder for static graphs is designed to reconstruct the original graphs from partially masked graphs. In particular, given a graph with node features represented by $ \bsX $ and an adjacency matrix denoted as $ \bsA $, we can apply random masking to obtain $ \bsX_m $ and $ \bsA_m$, encode them into a representation, and then decode the representation to reconstruct the original graph. Given a masking ratio $\alpha$, the masked node features $\bsX_m$ are constructed by substituting the randomly selected values with zeros or learnable parameters, and the masked adjacency matrix $\bsA_m$ is constructed by flipping randomly chosen subset of edges.  Either $ \bsX $ or $ \bsA $ or both can be masked before being passed to the encoder, depending on the self-supervised methodology. The masked features $ \bsX_m $ and $ \bsA_m $ are processed by an encoder $\mathcal{F}_{\texttt{enc}}$ (usually a \gls{gnn}) to be turned into a representation $\bsZ$, and then the representation $\bsZ$ is decoded via a decoder $\calF_{\texttt{dec}}$ to yield a reconstructed node features $\hat{\bsX}$:
\vspace{-4.5mm}

\begin{equation}
\bsZ = \mathcal{F}_{\texttt{enc}}(\bsX_m, \bsA_m), \quad \hat{\bsX} = \calF_{\texttt{dec}}(\bsZ).
\end{equation}
\vspace{-5.5mm}

The learning objective is to minimize the discrepancy between $\bsX$ and $\hat{\bsX}$, where the discrepancy can be \gls{mse}, \gls{bce}, or \gls{sce}. The decoder is usually constructed with \glspl{mlp}.
The adjacency matrix, based on an approach proposed in \citet{kipf2016variational}, can be reconstructed from the representation as $\hat{\bsA} = \texttt{sigmoid}(\bsZ\bsZ^\top)$.  The reconstruction loss between $\bsA$ and $\hat{\bsA}$ can also be included in the loss function to train the model.

\subsection{Constructing FC Network from fMRI Data}
\vspace{-1mm}

Following \citet{kim2021learning}, we construct dynamic graphs out of \gls{fc} networks in \gls{fmri} data by calculating the pairwise temporal correlation between the time series of different \glspl{roi}. Given a \gls{roi}-time series matrix $ \bsP \in \bbR^{N \times T_{\text{max}}} $, the \gls{fc} matrix $ \bsA(t) $ is defined as:
\begin{equation}
A_{ij}(t) = \frac{\text{Cov}(p_i(t), p_j(t))}{\sigma_{p_i}(t) \sigma_{p_j}(t)} \in \bbR^{N \times N}
\end{equation}
\vspace{-4mm}

To transform the correlation matrix into a binary adjacency matrix, we apply thresholding to the top 30-percentile of correlation values, marking them as connected edges. All other values are treated as unconnected, as described in \citet{kim2020understanding}.

\section{ST-MAE: Spatio-temporal Masked Autoencoder Frameworks}
\label{main:sec:method}
\vspace{-1mm}

\begin{figure}[t]
    \centering
        \includegraphics[width=\textwidth]{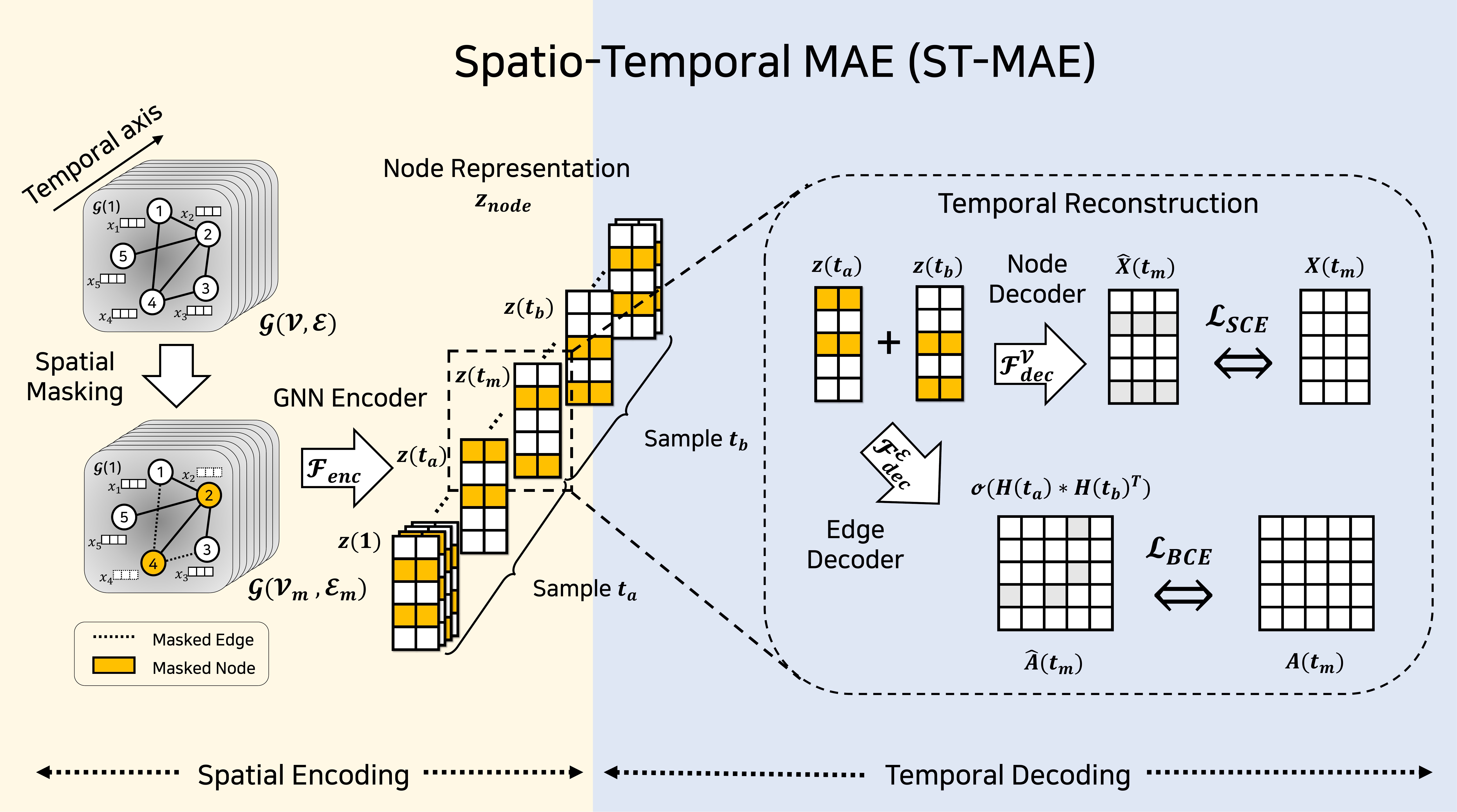}
    \caption{Spatio-Temporal Masked Autoencoder framework overview.}
    \label{fig:example}
\vspace{-3mm}
\end{figure}

In this study, we propose a generative \gls{ssl} approach for dynamic \gls{fc} of \gls{fmri} data. Unlike traditional static graph \gls{ssl} methods, our approach employs a \gls{gnn} encoder designed to capture knowledge in temporal graph data. To facilitate this, we use a masked autoencoding objective~\citep{he2021masked} to train an encoder for spatio-temporal graphs. These encoded representations are then leveraged to perform temporal reconstruction, where nodes and edges at an intermediate timestamp are reconstructed using encodings from different time points. This enables the model to integrate and learn from both the spatial and temporal dimensions of the graph.

\begin{algorithm}[t]
\setstretch{1.2}
\caption{Spatio-Temporal Masked Autoencoder (ST-MAE)}
\begin{algorithmic}
\STATE \textbf{Input:} Dynamic graph $ \mathcal{G}(t) $, Node features $ \bsX(t) $, Edge (\gls{fc}) matrix $ \bsA(t) $
\STATE \textbf{Output:} Spatial encoding $ \bsZ(t) $, Reconstructed node feature $ \hat{\bsX}(t_m) $, Edge (\gls{fc}) matrix $ \hat{\bsA}(t_m) $
\STATE Initialize \gls{gnn} encoder $ \mathcal{F}_{\texttt{enc}} $,   node decoder $ \mathcal{F}^{\mathcal{V}}_{\texttt{dec}}  $ and edge decoder $ \mathcal{F}^{\mathcal{E}}_{\texttt{dec}}  $
\FOR{each epoch}
\STATE $\calL_\texttt{spatial} \gets 0$ and $\calL_\texttt{temporal} \gets 0$.
\STATE Uniformly draw a subset $\calT \subseteq \{1,\dots, T\}$ of time-steps to apply masking.
    \FOR{$t \in \calT$}
    \STATE \texttt{/* Spatial reconstruction loss.*/}
        \STATE Mask nodes and edges in $ \bsX(t) $ and $ \bsA(t) $ to obtain $ \bsX_{\texttt{m}}(t) $ and $ \bsA_{\texttt{m}}(t) $
        \STATE Compute $\bsZ(t) \leftarrow \mathcal{F}_{\texttt{enc}}(\bsX_{\texttt{m}}(t), \bsA_{\texttt{m}}(t)) $.
        \STATE Compute $\hat{\bsX}(t) \gets \calF_\texttt{dec}^\calV(\bsW_{\text{sp}}\bsZ(t))$.
        \STATE Compute $\hat{\bsA}(t) = \texttt{sigmoid}(\bsH(t)\bsH(t)^\top)$, where $\bsH(t) = \calF_\texttt{dec}^\calV(\bsW_{\text{sp}}\bsZ(t))$.
        \STATE Compute the reconstruction loss and add it to $\calL_\texttt{spatial}$.

    \STATE \texttt{/* Temporal reconstruction loss.*/}
        \STATE Uniformly sample $(t_a, t_b)$ 
        from $\calS_{a,b}:=\{(t_a, t_b) | 1 \leq t_a < t < t_b \leq T\}$.
        \STATE Compute $\bsZ(t_a)$ and $\bsZ(t_b)$ with $\calF_\texttt{enc}$.
        \STATE Compute $\hat{\bsX}_{a,b}(t) = \calF_\texttt{dec}^\calV(\bsW_{\text{tp}}[\bsZ(t_a)\Vert \bsZ(t_b)])$. 
        \STATE Compute $\hat{\bsA}_{a,b}(t) = \frac{1}{2}\Big(\texttt{sigmoid}(\bsH(t_a)\bsH(t_b)^\top)+\texttt{sigmoid}(\bsH(t_b)\bsH(t_a)^\top)\Big)$.
        \STATE Compute the reconstruction loss and add it to $\calL_\texttt{temporal}$.
    \ENDFOR
    \STATE Compute the overall loss $\calL_\texttt{ST-MAE} = \calL_\texttt{spatial} + \calL_\texttt{temporal}$.
    \STATE Update the model parameters by taking the gradient descent step with $\calL_\texttt{ST-MAE}$.
\ENDFOR 
\end{algorithmic}
\label{alg:procedure}
\end{algorithm}

\subsection{Masked Autoencoding Objective for Capturing Spatial Patterns}
\vspace{-1mm}

Our framework is composed of a \gls{gnn} encoder $ \mathcal{F}_{\texttt{enc}} $ and two decoders $ \mathcal{F}^{\mathcal{V}}_{\texttt{dec}} $ and $ \mathcal{F}^{\mathcal{E}}_{\texttt{dec}} $ for reconstructing node features $ \bsX(t) $ and the adjacency matrix $ \bsA(t) $, respectively.

We first apply the masked autoencoding objective described in the previous section for individual time-steps $t$. Specifically, given a time-step $t$, we apply the masking to the node features and the adjacency matrix $(\bsX(t), \bsA(t))$ to obtain the masked versions $(\bsX_{\texttt{m}}(t), \bsA_{\texttt{m}}(t))$, and encode them to obtain a representation $\bsZ(t)$.
\begin{equation}
\bsZ(t) = \calF_\texttt{enc}(\bsX_{\texttt{m}}(t), \bsA_{\texttt{m}}(t)).    
\end{equation}

The node feature decoder $\calF_\texttt{dec}^\calV$ and the edge feature decoder $\calF_\texttt{dec}^\calE$ are then used to reconstruct
\begin{align}
\hat{\bsX}(t) = \calF_\texttt{dec}^\calV(\bsW_{\text{sp}}\bsZ(t)), \quad
\hat{\bsA}(t) = \texttt{sigmoid}(\bsH(t)\bsH(t)^\top)
\end{align}
where $\bsW_{\text{sp}} \in \bbR^{D \times D}$ is a learnable projection matrix and $\bsH(t) = \calF^\calE_\texttt{dec}(\bsW_{\text{sp}}\bsZ(t))$. 

At each training step, based on a pre-defined masking ratio, we pick a subset $\calT \subseteq \{1, \dots, T\}$ of time-steps and compute the reconstruction loss for those time-steps. We choose the \gls{sce} loss for the node reconstruction and the \gls{bce} loss for the adjacency reconstruction, constituting the spatial reconstruction loss,
\begin{equation}
\label{eq:stmae_recon}
    \calL_\texttt{spatial} = \sum_{t\in \calT} \Big(\calL_\texttt{sce}(\bsX(t), \hat{\bsX}(t)) + \calL_\texttt{bce}(\bsA(t), \hat{\bsA}(t))\Big).
\end{equation}

\subsection{Temporal Reconstruction Objective}

To further encourage the encoder to capture the temporal dynamics in graphs, we employ the additional task for our self-supervised learning framework which is to predict a graph at a  time step $t$ based on the representations computed from the graphs at nearby time steps. More specifically, for $t \in \calT$, we first draw two timesteps $t_a$ and $t_b$ uniformly from $\calS_{a,b}:=\{(t_a, t_b) | 1 \leq t_a < t < t_b \leq T\}$.
The task is to reconstruct $(\hat{\bsX}_{a,b}(t), \hat{\bsA}_{a,b}(t))$ based on the representations $\bsZ(t_a)$ and $\bsZ(t_b)$, not based on the representation computed from the masked version of the graph $(\bsX(t), \bsA(t))$ as before. The node feature decoder $\calF_\texttt{dec}^\calV$  reconstructs the node feature $\hat{\bsX}_{a,b}(t)$ based on two representations,
\begin{equation}
    \hat{\bsX}_{a,b}(t) = \calF^\calV_\texttt{dec}(\bsW_{\text{tp}}[\bsZ(t_a)\Vert \bsZ(t_b)])
\end{equation}
where $\bW_{\text{tp}} \in \bbR^{2D \times D}$ is a learnable projection matrix. 
The adjacency matrix is reconstructed similarly, but using two representations $\bsZ(t_a)$ and $\bsZ(t_b)$,
\begin{align}
    &\bsH(t_a) = \calF_\texttt{dec}^\calE(\bsW_{\text{sp}}\bsZ(t_a)), \,\, \bsH(t_b) = \calF_\texttt{dec}^\calE(\bsW_{\text{sp}}\bsZ(t_b)), \\%\,\, 
    \hat{\bsA}_{a,b}(t) =&\,\frac{1}{2}\Big(\texttt{sigmoid}(\bsH(t_a)\bsH(t_b)^\top) + \texttt{sigmoid}(\bsH(t_b)\bsH(t_a)^\top)\Big).
\end{align}
Then we compute the temporal reconstruction loss similar to the spatial reconstruction loss as,
\begin{equation}
\calL_\texttt{temporal} =\sum_{t\in \calT} \Big(\calL_\texttt{sce}(\hat{\bsX}_{a,b}(t), \bsX(t)) + \calL_\texttt{bce}(\hat{\bsA}_{a,b}(t), \bsA(t))\Big).    
\end{equation}

\subsection{Overall Training Pipeline}
At each step, we compute the spatial reconstruction loss $\calL_\texttt{spatial}$ and the temporal reconstruction loss $\calL_\texttt{temporal}$. The overall loss function $\calL_\texttt{ST-MAE}$ is defined as the sum of the two objectives. 
\begin{equation}
\calL_\texttt{ST-MAE} = \calL_\texttt{spatial} + \calL_\texttt{temporal}
\end{equation}
We call our self-supervised learning framework based on masked autoencoder the Spatio-Temporal Masked Autoencoder (ST-MAE) for dynamic graphs. \cref{alg:procedure} summarizes the overall training pipeline of ST-MAE.

\section{Experiments}
\label{main:sec:experiments}

\textbf{Datasets.}
We compare our proposed method with several state-of-the-art \gls{ssl} methods on a collection of publicly available resting-state \gls{fmri} datasets including both static and dynamic circumstances. We preprocess \gls{fmri} data into dynamic graphs with \gls{fc} of 400 \glspl{roi}. As UKB \cite{sudlow2015uk} consists of 40,913 samples, which is one of the largest public \gls{fmri} datasets, we use it for pre-training. Then, we present downstream findings on six datasets: ABCD \cite{casey2018adolescent}, HCP \cite{wu20171200}, HCP-A \cite{bookheimer2019lifespan}, HCP-D \cite{somerville2018lifespan}, ABIDE \cite{heinsfeld2018identification}, and ADHD200 \cite{brown2012adhd}. Graph statistics under dynamic settings are in \cref{main:tab:dataset_info}. Please refer to the details of the datasets and baselines in Appendix \textcolor{linkcolor}{A}.

\begin{table}[t]

\caption{Statistics of dynamic graphs in \gls{fmri} datasets. The variables represent the following; $\left| G \right|$: number of graphs, $\left| N \right|$: number of nodes, $\left| E \right|$: number of edges, $d_{max}$: the maximum degree of nodes in each dataset, $d_{avg}$: average degree of nodes in each dataset, $K$: global clustering coefficient.}
\centering
\begin{tabular}{l|ccccccc}

\toprule
Dataset %
& $\left| G \right|$ & $\left| N \right|_{avg}$ & $\left| E \right|_{avg}$ & $d_{max}$ & $d_{avg}$ & $K$ \\
\midrule
    
UKB             & 1,145,564 & 400 & 23,800 & 264 & 119 & 0.662 \\ %
ABCD            & 191,331 & 400 & 23,800 & 285 & 119 & 0.663 \\ %
HCP             & 78,696 & 400 & 23,800 & 281 & 119 & 0.601 \\ %
HCP-A           & 19,548 & 400 & 23,800 & 287 & 119 & 0.644 \\ %
HCP-D           & 17,064 & 400 & 23,800 & 262 & 119 & 0.633 \\ %
ABIDE           & 83,096 & 400 & 23,800 & 278 & 119 & 0.616 \\ %
ADHD200         & 36,126 & 400 & 23,800 & 257 & 119 & 0.592 \\ %
\bottomrule
\end{tabular}
\label{main:tab:dataset_info}
\end{table}

\subsection{Experimental Details}
To construct dynamic graphs, we employed a window size and stride of 50 and 16, respectively, for the UKB, ABCD, HCP, HCP-A, and HCP-D datasets. For the ABIDE and ADHD200 datasets, we used values of 16 and 3. Additionally, we followed a procedure akin to that described in \citet{kim2021learning}, wherein each batch containing \gls{roi}-timestamps of fixed length sampled randomly per dataset.

For the baseline of our experiment, we employed a 4-layer \gls{gin}~\citep{xu2018powerful} as \gls{gnn} encoder. Following \citet{kim2021learning}, to obtain the graph representation, We used SERO as the readout function and leveraged a jumping knowledge network~\citep{xu2018representation} architecture, which concatenates dynamic graph representations across layers. 

For the pre-training of the \gls{gnn} encoder, we used the UKB dataset, which consists of 40,913 samples. We evaluated the downstream performance for tasks such as gender classification and age regression on a diverse set of public \gls{fmri} datasets, including ABCD, HCP, HCP-A, HCP-D, ABIDE, and ADHD200. Furthermore, to assess potential improvements in clinical classification, we tested psychiatric disorder classification performance on the ABIDE and ADHD200 datasets. We use Adam optimizer with a learning rate of 0.0005 and a weight decay of 0.0001. During pre-training, we used a cosine decay learning rate scheduler, while for fine-tuning, a one-cycle scheduler was employed. Specifically, the learning rate increased gradually to 0.001 during the initial 20\% of the training epochs and then decreased to 5.0 $\times$ $10^{-7}$. Our approach was consistently trained with a batch size of 32. All experiments were conducted on an NVIDIA GeForce RTX 3090. The fine-tuning performance was averaged over 5-fold cross-validation.

\subsection{Downstream-task Performance}

\begin{table}[t]
\caption{Results for gender classification tasks across \gls{fmri} datasets. Scores represent the \gls{auroc}. 
}
\begin{adjustbox}{width=\linewidth, totalheight=\textheight, keepaspectratio}
\begin{tabular}{c|c|c|cccccc|c}

\toprule
Type of FC & Methods & Train Type & ABCD & HCP &  HCPA &  HCPD & ABIDE &  ADHD200 & Rank \\
\midrule
\multirow{6}{*}{Static}
  & Baseline & Supervised & $\underline{83.75}$ & 86.31 & 68.36 & 65.13 & 68.81 & 61.60 & 6.83 \\
\cline{2-9}
  & DGI~\citep{velivckovic2018deep} & \multirow{2}{*}{Contrastive SSL} & 73.80 & 87.02 & 70.15 & 68.20 & 67.89 & 62.17 & 6.50 \\
  & SimGRACE~\citep{xia2022simgrace} &  & 73.93 & 87.40 & 69.60 & 66.77 & 70.47 & 65.08 & 5.17 \\
\cline{2-9}
  & GAE~\citep{kipf2016variational} & \multirow{3}{*}{Generative SSL} & 73.45 & 87.31 & 70.66 & 68.39 & 69.90 & 62.88 & 5.50 \\
  & VGAE~\citep{kipf2016variational} &  & 72.84 & 87.05 & 68.28 & 65.09 & 71.31 & 64.14 & 6.83 \\
  & GraphMAE~\citep{hou2022graphmae} &  & 72.79 & 87.77 & 66.87 & 66.42 & 66.98 & 61.48 & 7.83 \\
\midrule
\multirow{4}{*}{Dynamic}
  & Baseline & \multirow{1}{*}{Supervised} & $\mathbf{85.06}$ & $\underline{93.10}$ & $\underline{84.24}$ & $\underline{73.19}$ & $\underline{73.91}$ & $\underline{72.12}$ & $\underline{1.83}$ \\
\cline{2-9}
  & ST-DGI~\cite{opolka2019spatio} & \multirow{1}{*}{Contrastive SSL} & 83.14 & 92.50 & 82.73 & 70.85 & 72.00 & 65.69 & 3.17 \\
\cline{2-9}
  & ST-MAE (Ours) & \multirow{1}{*}{Generative SSL} & 83.15 & $\mathbf{93.58}$ & $\mathbf{86.32}$ & $\mathbf{74.92}$ & $\mathbf{77.89}$ & $\mathbf{72.68}$ & $\mathbf{1.33}$ \\
\bottomrule
\end{tabular}

\end{adjustbox}
\vspace{-3mm}
\label{main:tab:gender}
\end{table}

\begin{table}[t]
\caption{Results for age regression tasks across \gls{fmri} datasets. Scores represent the \gls{mae}.
}
\begin{adjustbox}{width=\linewidth, totalheight=\textheight, keepaspectratio}
\begin{tabular}{c|c|c|cccccc|c}

\toprule
Type of FC & Methods & Train Type & ABCD & HCP &  HCPA &  HCPD & ABIDE & ADHD200 & Rank \\
\midrule

\multirow{6}{*}{Static}
  & Baseline & Supervised & $\mathbf{0.51}$ & 3.11 & 9.44 & 2.51 & 4.39 & 2.07 & 6.67 \\
\cline{2-9}
  & DGI~\citep{velivckovic2018deep} & \multirow{2}{*}{Contrastive SSL} & $\underline{0.54}$ & 3.12 & 9.38 & 2.50 & 4.27 & 2.03 & 5.33 \\
  & SimGRACE~\citep{xia2022simgrace} &  & $\underline{0.54}$ & 3.09 & 9.48 & 2.35 & 4.28 & 1.97 & 5.00 \\
\cline{2-9}
  & GAE~\citep{kipf2016variational} & \multirow{3}{*}{Generative SSL} & $\underline{0.54}$ & 3.12 & 9.29 & 2.42 & 4.37 & 2.06 & 6.00 \\
  & VGAE~\citep{kipf2016variational} &  & $\underline{0.54}$ & 3.13 & 9.40 & 2.39 & 4.26 & 2.07 & 6.33 \\
  & GraphMAE~\citep{hou2022graphmae} &  & $\underline{0.54}$ & 3.08 & 9.43 & 2.48 & 4.41 & 2.05 & 6.50 \\
\midrule
\multirow{3}{*}{Dynamic}
  & Baseline & \multirow{1}{*}{Supervised} & 0.55 & $\mathbf{2.74}$ & $\underline{8.39}$ & 2.16 & $\mathbf{4.12}$ & 1.97 & 3.50 \\
\cline{2-9}
  & ST-DGI~\cite{opolka2019spatio} & \multirow{1}{*}{Contrastive SSL} & $\underline{0.54}$ & 2.84 & $\mathbf{7.93}$ & $\underline{2.15}$ & 4.18 & $\underline{1.93}$ & $\underline{3.00}$ \\
\cline{2-9}
  & ST-MAE (Ours) & \multirow{1}{*}{Generative SSL} & $\underline{0.54}$ & $\underline{2.82}$ & $\mathbf{7.93}$ & $\mathbf{2.06}$ & $\underline{4.13}$ & $\mathbf{1.86}$ & $\mathbf{2.67}$ \\
\bottomrule
\end{tabular}
\end{adjustbox}
\label{main:tab:age}
\end{table}

\begin{table}[t]
\caption{Results for psychiatric diagnosis classification tasks on ABIDE and ADHD200 datasets. 
}
\begin{adjustbox}{width=\linewidth, totalheight=\textheight, keepaspectratio}
\begin{tabular}{c|c|c|cccc|c}
\toprule

\multirow{2}{*}{Type of FC} & \multirow{2}{*}{Methods} & \multirow{2}{*}{Train Type}& \multicolumn{2}{c}{ABIDE} & \multicolumn{2}{c}{ADHD200} & \multirow{2}{*}{Rank} \\
 & & & Acc. $(\uparrow)$ & AUROC $(\uparrow)$ & Acc. $(\uparrow)$ & AUROC $(\uparrow)$ & \\
\midrule

\multirow{6}{*}{Static}
  & Baseline & Supervised & 58.94 & 63.78 & 49.47 & 55.74 & 7.00 \\
\cline{2-8}
  & DGI~\citep{velivckovic2018deep} & \multirow{2}{*}{Contrastive SSL} & 60.52 & 64.44 & 49.17 & 54.94 & 7.25 \\
  & SimGRACE~\citep{xia2022simgrace} &  & 60.97 & 66.14 & 45.88 & 54.50 & 7.00 \\
\cline{2-8}
  & GAE~\citep{kipf2016variational} & \multirow{3}{*}{Generative SSL} & 61.09 & 65.14 & 48.57 & 55.96 & 5.50 \\
  & VGAE~\citep{kipf2016variational} &  & 62.44 & 65.04 & 50.67 & $\underline{58.50}$ & 4.00 \\
  & GraphMAE~\citep{hou2022graphmae} &  & 61.65 & 64.46 & 52.01 & 55.37 & 5.25 \\
\midrule
\multirow{3}{*}{Dynamic}
  & Baseline & Supervised & $\underline{63.01}$ & $\underline{67.58}$ & $\underline{52.47}$ & 58.27 & $\underline{2.25}$ \\
\cline{2-8}
  & ST-DGI~\cite{opolka2019spatio} & \multirow{1}{*}{Contrastive SSL} & 62.79 & 67.03 & 48.27 & 54.47 & 5.75 \\
\cline{2-8}
  & ST-MAE (Ours) & \multirow{1}{*}{Generative SSL} & $\mathbf{64.48}$ & $\mathbf{69.03}$ & $\mathbf{53.07}$ & $\mathbf{59.35}$ & $\mathbf{1.00}$ \\

\bottomrule
\end{tabular}
\end{adjustbox}
\label{main:tab:diagnosis}
\end{table}

We evaluated the performance of ST-MAE using multiple publicly available \gls{fmri} datasets, with particular emphasis on gender classification, age regression, and psychiatric diagnosis classification tasks. The empirical results reported in \cref{main:tab:gender}, \cref{main:tab:age}, and \cref{main:tab:diagnosis} clearly show that our method consistently outperforms both self-supervised and supervised baselines across all tasks.

For gender classification in \cref{main:tab:gender}, ST-MAE achieved the highest \gls{auroc} scores, particularly excelling in dynamic FC with an \gls{auroc} of 77.89 on the ABIDE dataset. Similarly, in the age regression task in \cref{main:tab:age}, ST-MAE demonstrated superiority by achieving the lowest \gls{mae} in the HCP-D and ADHD200 datasets. Moreover, in psychiatric diagnosis classification in \cref{main:tab:diagnosis}, particularly where labeled data are scarce, ST-MAE outperforms other models on the ABIDE and ADHD200 datasets.

These results validate the effectiveness of ST-MAE in capturing both spatial and temporal dynamics, while also highlighting its broad applicability and robustness in real-world scenarios. Importantly, by leveraging \gls{ssl}, ST-MAE addresses the challenge of limited labeled data, making it particularly impactful for advancing research in neuropsychiatric disorders and other healthcare applications reliant on \gls{fmri} data analysis.

\subsection{Ablation Study}
We aimed to take full advantage of the large number of unlabeled fMRI data to develop a useful fMRI representation through \gls{ssl} for downstream tasks with relatively limited data.
To demonstrate the effectiveness of ST-MAE, we conducted an ablation study on the number of data for \gls{ssl} and labeled data ratio for downstream task, and reconsruction strategies.

\begin{figure}[th]
    \centering

    \begin{minipage}{.32\textwidth}
        \centering
        \includegraphics[width=\linewidth]{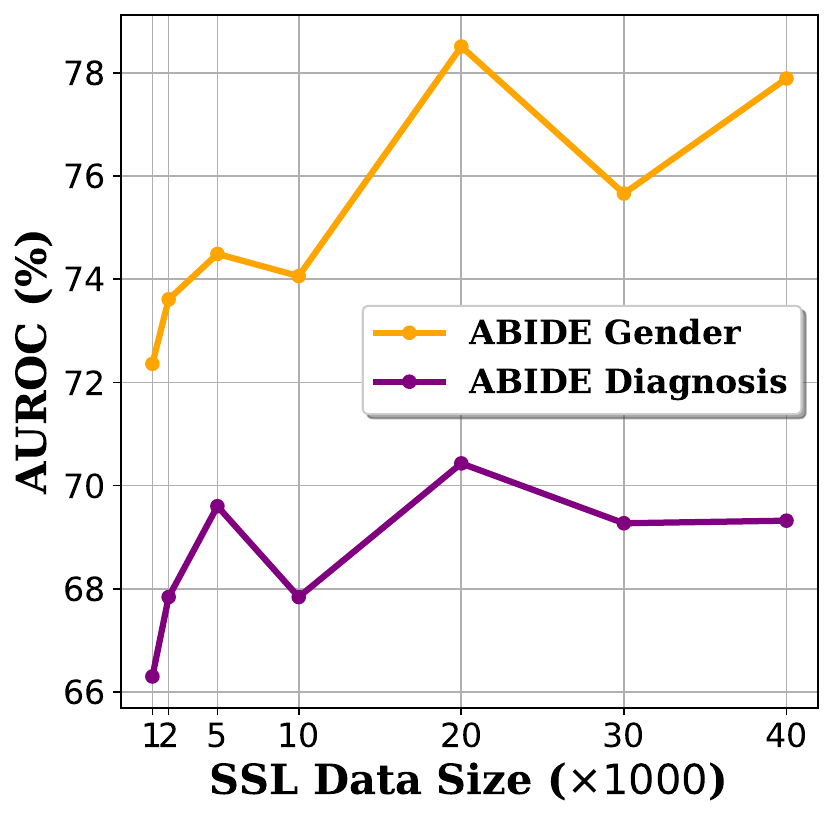}
        \caption{The effect of the number of data for \gls{ssl}.}
        \label{fig:ssl_data_size}
    \end{minipage}
    \hfill
    \begin{minipage}{.32\textwidth}
        \centering
        \includegraphics[width=\linewidth]{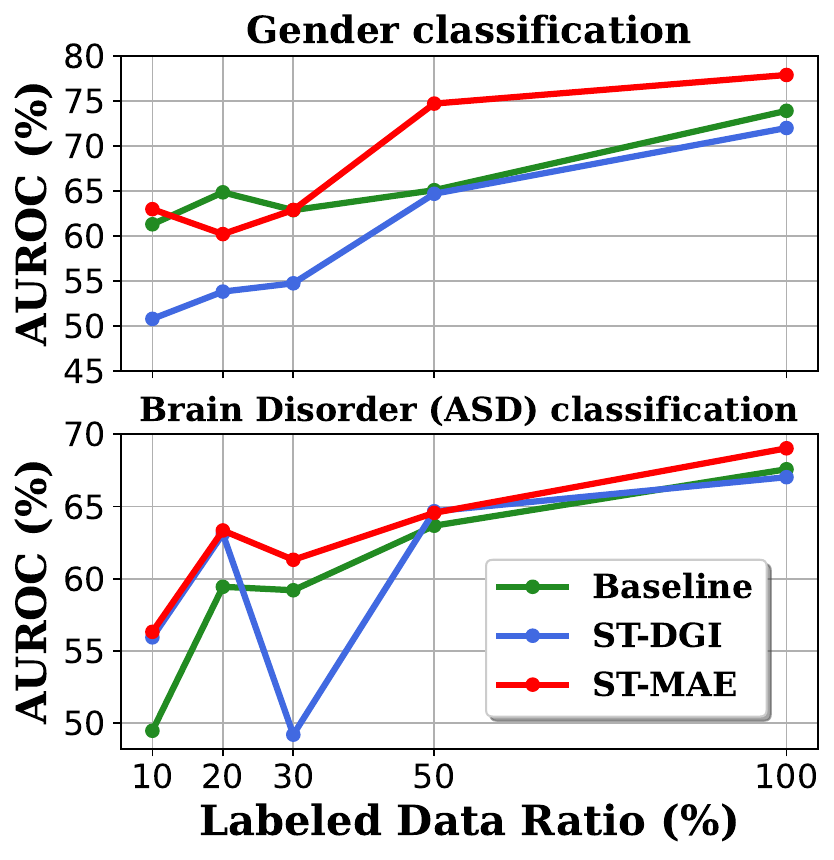}
        \vspace{-18px}
        \caption{ABIDE classification results on limited data.}
        \label{fig:limited_data}
    \end{minipage}
    \hfill
    \begin{minipage}{.32\textwidth}
        \centering
        \includegraphics[width=\linewidth]{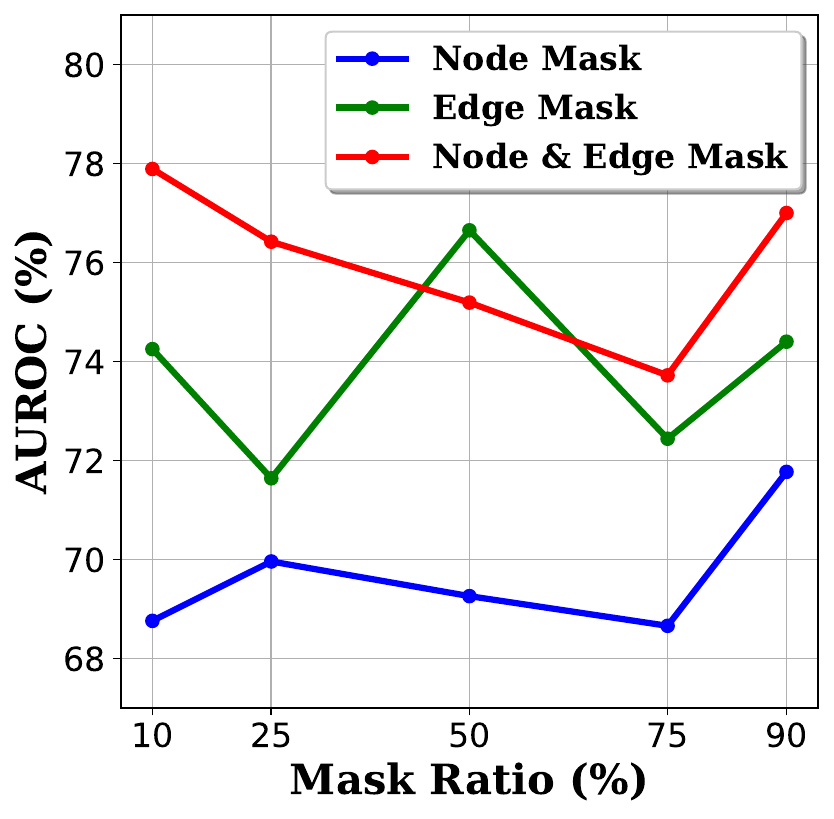}
        \caption{The ablation result of mask ratio on ABIDE dataset.}
        \label{fig:ablation_maskratio}
    \end{minipage}
    
\end{figure}

\subsubsection{Effectiveness of Large-scale fMRI Datasets}
 We examined the impact of the amount of UKB data used for \gls{ssl} on downstream performance, using gender classification on the ABIDE dataset as a case study.
 As shown in Figure \ref{fig:ssl_data_size}, we confirmed our intuition that performance increases as the amount of data used for \gls{ssl} increases. 
 This confirms that it is possible to learn a meaningful fMRI representation from large scale fMRI data though \gls{ssl}.

\subsubsection{Effectiveness for Limited Data}

In scenarios with a limited number of labels, we reduced the percentage of labeled data used for downstream training to see if ST-MAE could achieve better performance with less data. In Figure \ref{fig:limited_data}, we observe that the model performing \gls{ssl} with ST-MAE achieves better performance even when trained using less data, suggesting that it provided a more useful starting point for downstream tasks.

\subsubsection{Ablation of Masking Ratio}

To see how the method used for reconstruction and the masking ratio affect the performance of the downstream task, we trained nodes and edges while varying the masking ratio and measured the performance of gender classification on ABIDE dataset. In Figure \ref{fig:ablation_maskratio}, we can see that using both nodes and edges for restoration is more effective than learning them separately, and the performance difference due to the masking ratio varies in a manner similar to the performance difference of the individual reconstruction targets. Since performance can vary depending on the masking ratio, it is important to specify the appropriate masking ratio according to the task.

\subsubsection{Ablation of Reconstruction Criterion}
\begin{wraptable}{r}{6.5cm}
\vspace{-4.5mm}
\caption{Ablation results of reconsruction criterion on ABIDE and ADHD200 datasets}
\begin{adjustbox}{width=\linewidth, totalheight=\textheight, keepaspectratio}
\begin{tabular}{cc|cc|cc}
\toprule
\multicolumn{2}{c}{Criterion} & \multicolumn{2}{c}{ABIDE} & \multicolumn{2}{c}{ADHD200} \\
\hline
Node & Edge & gender & diagnosis & gender & diagnosis \\
\midrule
MSE & MSE & 72.21 & 64.26 & 64.88 & 54.06 \\
MSE & BCE & 74.20 & 63.57 & 69.25 & 54.22 \\
SCE & MSE & 74.96 & 63.58 & 66.49 & 55.10 \\
SCE & BCE & \textbf{77.89} & \textbf{65.05} & \textbf{70.29} & \textbf{55.27} \\
\bottomrule
\end{tabular}
\end{adjustbox}

\label{tab:criterion_ablation}
\end{wraptable}

We compared the reconstruction criterion used in ST-MAE with different criteria for each of the node and edge reconstructions. For node reconstruction, we used \gls{mse} and \gls{sce}, and for edge reconstruction, we used \gls{mse} and \gls{bce} to compare the effectiveness of each combination. As shown in Table \ref{tab:criterion_ablation}, we found the best combination when using \gls{sce} as the node restoration criterion and \gls{bce} as the edge restoration criterion, and this combination was incorporated into our ST-MAE framework.

\section{Related Works}
\label{main:sec:relatedworks}

\subsection{Self-supervised Learning on Static Graphs}
\gls{ssl} on static graphs has emerged as a compelling approach to extract useful representations from graph-structured data without requiring explicit labels. These methods are generally classified into two categories: contrastive \gls{ssl} and generative \gls{ssl}. Both approaches aim to generate informative node and edge features that are useful for a variety of downstream tasks, such as node classification, link prediction, and graph classification.

\textbf{Contrastive Self-supervised Learning}
Contrastive \gls{ssl} techniques in graphs aim to learn embeddings by maximizing the similarity between closely related nodes while minimizing the similarity between unrelated nodes. DGI~\citep{velivckovic2018deep} was a foundational work that introduced the concept of maximizing mutual information between local patches and the entire graph. GCL~\citep{you2020graph} extended this by leveraging graph augmentations to create positive pairs. Though these methods offer better generalization capabilities, they come at the cost of computational efficiency. To mitigate this, SimGRACE~\citep{xia2022simgrace} provided a simplified approach that omits the need for complex data augmentations, and SimGCL~\citep{yu2022graph} introduced the use of InfoNCE loss for generating contrastive samples.

\textbf{Generative Self-supervised Learning}
Generative \gls{ssl} in graphs primarily focuses on reconstructing the original graph or its features from partially masked or perturbed node or edge features. VGAE~\citep{kipf2016variational}, a pioneering work in generative \gls{ssl}, proposed a method for reconstructing a graph's adjacency matrix using node representations. It employed \gls{vae} for unsupervised learning in graph-structured data, achieving effective performance in link prediction tasks.
GraphMAE~\citep{hou2022graphmae}, as one of the earliest works in this area, concentrated on the reconstruction of node features and demonstrated superior performance in node and graph classification tasks over traditional contrastive self-supervised learning methods, thanks to its simpler restoration techniques. Building on this, GraphMAE2~\citep{hou2023graphmae2} introduced multi-view random masking and regularization, further enhancing generalization performance. However, these methods primarily focus on static graphs and do not consider learning the temporal dynamics inherent in dynamic graphs.

\subsection{Self-supervised Learning on Dynamic Graphs}
\gls{ssl} techniques for dynamic graphs are relatively less explored, especially in the medical domain. These methods aim to capture the evolving nature of graphs, emphasizing the temporal relationships among nodes in addition to the spatial structure. Some pioneering work has been done in non-medical sectors like traffic flow prediction~\citep{opolka2019spatio, zhang2023spatial, ji2023spatio}. For instance, Ti-MAE~\citep{opolka2019spatio} has shown how generative \gls{ssl} can be effective for time-series graph data, particularly in overcoming distribution shift issues commonly seen in contrastive approaches.

\subsection{Deep Neural Networks on Spatio-Temporal Graphs}
Deep learning on spatio-temporal graphs is a burgeoning field that aims to capture both the spatial relationships and temporal dynamics in graph-structured data. STAGIN~\citep{kim2021learning} was a seminal work that successfully integrated both spatial and temporal aspects, setting a new performance benchmark across multiple tasks. This serves as our baseline for \gls{ssl} on spatio-temporal graphs. Following this, NeuroGraph~\citep{said2023neurograph} introduced a benchmark dataset and demonstrated performance improvements by utilizing sparser graphs and a larger number of \glspl{roi}.

\section{Conclusion}
\label{main:sec:conclusion}

In this study, we presented Spatio-Temporal Masked AutoEncoder (ST-MAE), a \gls{ssl} framework tailored for \gls{fmri} dynamic graphs. Our method has shown robust and superior performance in various downstream tasks, ranging from gender classification to psychiatric diagnosis classifcation. Our work contributes to both the \gls{fmri} research community and the broader field of \gls{ssl}, especially in settings where labeled data are limited. The findings affirm that ST-MAE excels not only in capturing spatio-temporal dynamics but also in its adaptability for a wide range of applications. We believe this work opens up new possibilities for more advanced analytics in multiple domains.

\begin{ack}
This work was partly supported by Basic Science Research Program through the National Research Foundation of Korea(NRF) funded by the Ministry of Education (NRF-2022R1I1A1A01069589), the National Research Foundation of Korea(NRF) grant funded by the Korea government(MSIT) (NRF-2021M3E5D9025030) and Institute of Information \& communications Technology Planning \& Evaluation (IITP) grant funded by the Korea government(MSIT) (No.2019-0-00075, Artificial Intelligence Graduate School Program (KAIST)).
\end{ack}

\clearpage
\newpage
\appendix
\section{Datasets}
\label{app:sec:dataset}

\begin{itemize}

    \item \textbf{UK Biobank (UKB)}~\citep{sudlow2015uk}: Consisting of 40,913 samples, the UK Biobank dataset stands as one of the most comprehensive fMRI datasets available. The dataset includes extensive demographic information, such as gender and age (between 40 and 70 years), which are valuable for various pre-training tasks.

    \item \textbf{Adolescent Brain Cognitive Development (ABCD)}~\citep{casey2018adolescent}: This dataset comprises 9,111 samples from children and adolescents aged 9 to 11 years, focusing on their development. It includes demographic information on gender and age, useful for developmental studies and can be utilized alongside the UKB dataset for pre-training.

    \item \textbf{Human Connectome Project (HCP) Young Adults}~\citep{wu20171200}: The HCP Young Adults dataset includes 1,093 samples from participants aged 22 to 37 years, providing a valuable resource for studying brain connectivity in young adults.

    \item \textbf{Human Connectome Project (HCP) Aging}~\citep{bookheimer2019lifespan}: The HCP-A dataset, with 724 samples, focuses on older adults aged 36 to 90 years, offering insights into brain changes and development in this age group.

    \item \textbf{Human Connectome Project (HCP) Development}~\citep{somerville2018lifespan}: The HCP-D dataset, consisting of 632 samples, targets the developmental stages of children and adolescents, encompassing ages from 8 to 21 years. It provides gender and age data for detailed developmental analyses.
    
    \item \textbf{Autism Brain Imaging Data Exchange (ABIDE)}~\citep{craddock2013neuro}: The ABIDE dataset includes 884 clinical samples and provides Autism Spectrum Disorder (ASD) labels, making it useful for benchmarking psychiatric diagnosis classification tasks.
    
    \item \textbf{ADHD200}~\citep{brown2012adhd}: This dataset includes 669 clinical samples and contains labels for Normal and ADHD conditions, serving as a useful resource for benchmarking psychiatric diagnosis classification.
\end{itemize}

\section{Baseline Graph Self-supervised Methods}
\label{app:sec:baseline}

\begin{itemize}
    \item \textbf{Deep Graph Infomax (DGI)}~\citep{velivckovic2018deep}: DGI aims to maximize the mutual information between node representations and global graph representations. A discriminator is trained to differentiate between the original graph and a permuted version, thereby learning meaningful node and graph representations.
    
    \item \textbf{Graph Auto-Encoder (GAE)}~\citep{kipf2016variational}: GAE employs an autoencoder architecture to reconstruct the original graph from node representation. The model learns to infer node features with adjacency matrix $\bsA$ and uses them to reconstruct the original links of graph.
    
    \item \textbf{Variational Graph Auto-Encoder (VGAE)}~\citep{kipf2016variational}: VGAE extends GAE by introducing stochasticity in the encoder layer. The encoder outputs the mean and standard deviation, from which node representations are sampled. These sampled representations are then used to reconstruct the original graph. The reconstruction is given by \( \hat{\bsA} = \sigma(\bsZ\bsZ^T) \), where \( \bsZ = \text{GCN}(\bsX, \bsA) \).
    
    \item \textbf{SimGRACE}~\citep{xia2022simgrace}: Unlike traditional \gls{gcl} methods that use graph augmentations to create multiple views, SimGRACE perturbs the model weights to generate different views. This approach eliminates the need for dataset-specific augmentations, making it a more universally applicable method~\citep{xia2022simgrace}.
    
    \item \textbf{Spatio-Temporal Deep Graph Infomax (ST-DGI)}~\citep{opolka2019spatio}: ST-DGI extends DGI to spatio-temporal graphs. It trains a discriminator to differentiate between node features at different time steps, thus capturing both spatial and temporal dynamics of the graph.
    
    \item \textbf{Graph Masked AutoEncoder (GraphMAE)}~\citep{hou2022graphmae}: GraphMAE focuses on masked node feature reconstruction rather than edge reconstruction. Its successor, GraphMAE2, further enhances the model by introducing additional regularization techniques for better performance~\citep{hou2023graphmae2}.
\end{itemize}

\end{document}